\documentclass[10pt,journal,compsoc]{IEEEtran}
\usepackage{multirow}
\usepackage{comment}
\usepackage{bm} 
\usepackage[table]{xcolor}
\usepackage{graphicx}
\usepackage{booktabs} 

\newcounter{ablStdModel} 
\newcommand{\ablStdModel}{\refstepcounter{ablStdModel}\theablStdModel}

\newcommand{\keypoint}[1]{\vspace{0.1cm}\noindent\textbf{#1}\quad}

\newcommand{\red}[1]{\textcolor{red}{#1}}
\newcommand{\blue}[1]{\textcolor{blue}{#1}}
\newcommand{\green}[1]{\textcolor{green}{#1}}
\newcommand{\tableCellHeight}{1}

\ifCLASSOPTIONcompsoc
  \usepackage[nocompress]{cite}
\else
  \usepackage{cite}
\fi
\ifCLASSINFOpdf
\else
\fi
\usepackage{amsmath}
\usepackage{amssymb}
\usepackage{algorithmic}
\usepackage{url}

\begin{document}
%
\title{Learning Generalisable Omni-Scale Representations for Person Re-Identification}
%
%
%
%

\author{Kaiyang Zhou,
Yongxin Yang,
Andrea Cavallaro,
and Tao Xiang

\IEEEcompsocitemizethanks{
\IEEEcompsocthanksitem K.~Zhou is with Nanyang Technological University, Singapore. E-mail: \{kaiyang.zhou\}@ntu.edu.sg

\IEEEcompsocthanksitem Y.~Yang and T.~Xiang are with the University of Surrey, Guildford, GU2 7XH, UK. E-mail: \{k.zhou, y.yang, t.xiang\}@surrey.ac.uk

\IEEEcompsocthanksitem A.~Cavallaro is with Queen Mary University of London, London, E1 4NS, UK. E-mail: a.cavallaro@qmul.ac.uk
}
}

%
%

\markboth{IEEE Transactions on Pattern Analysis and Machine Intelligence}
{Shell \MakeLowercase{\textit{et al.}}: Bare Demo of IEEEtran.cls for Computer Society Journals}
%



\IEEEtitleabstractindextext{%
\begin{abstract}
An effective person re-identification (re-ID) model should learn feature representations that are both discriminative, for distinguishing similar-looking people, and generalisable, for deployment across datasets without any adaptation. In this paper, we develop novel CNN architectures to address both challenges. First, we present a re-ID CNN termed omni-scale network (OSNet) to learn features that not only capture different spatial scales but also encapsulate a synergistic combination of multiple scales, namely omni-scale features. The basic building block consists of multiple convolutional streams, each detecting features at a certain scale. For omni-scale feature learning, a unified aggregation gate is introduced to dynamically fuse multi-scale features with channel-wise weights. OSNet is lightweight as its building blocks comprise factorised convolutions. Second, to improve generalisable feature learning, we introduce instance normalisation (IN) layers into OSNet to cope with cross-dataset discrepancies. Further, to determine the optimal placements of these IN layers in the architecture, we formulate an efficient differentiable architecture search algorithm. Extensive experiments show that, in the conventional same-dataset setting, OSNet achieves state-of-the-art performance, despite being much smaller than existing re-ID models. In the more challenging yet practical cross-dataset setting, OSNet beats most recent unsupervised domain adaptation methods without using any target data. Our code and models are released at \texttt{https://github.com/KaiyangZhou/deep-person-reid}.
\end{abstract}

}

\maketitle

\IEEEdisplaynontitleabstractindextext

%
\IEEEpeerreviewmaketitle

\section{Introduction} \label{sec:intro}

Person re-identification (re-ID), as a fine-grained instance recognition problem, aims to match people across non-overlapping camera views. With the development of deep learning technology, recent research in person re-ID has shifted from tedious feature  engineering~\cite{liao2015person,matsukawa2016hierarchical} to end-to-end feature representation learning with deep neural networks~\cite{li2014deepreid,ahmed2015improved,li2018harmonious,chang2018multi}, especially convolutional neural networks (CNNs).

\begin{figure}[t]
\centering
\includegraphics[width=0.99\columnwidth]{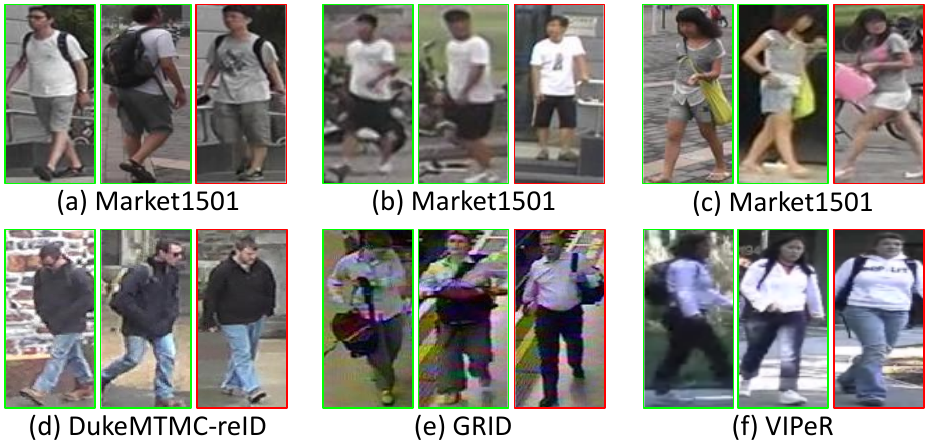}
\caption{Example images from four person re-ID datasets showing that discriminative and generalisable features are essential for re-ID. Each sub-figure contains, from left to right, a query image, a true match, and a false match (distractor).}
\label{fig:reid_examples}
\end{figure}

Though the re-ID performance has been improved significantly thanks to end-to-end representation learning with CNNs, two problems remain unsolved. They hinder large-scale deployment of re-ID models in real-world applications. The first problem is \emph{discriminative} feature learning. As an instance recognition task, re-identifying people under disjoint camera views needs to overcome both intra-class variations and inter-class ambiguity. For instance, in Fig.~\ref{fig:reid_examples}(a) the view change from front to back across cameras brings large appearance changes in the backpack region, making person matching a challenging task. Moreover, from a distance as typical in video surveillance scenes, people can look incredibly similar, as exemplified by the false matches in Fig.~\ref{fig:reid_examples}. This requires the re-ID features to capture fine-grained details for distinguishing people of similar appearances (e.g., the sun glasses in Fig.~\ref{fig:reid_examples}(d)).

The second problem is \emph{generalisable} feature learning. Due to intrinsic domain gaps between re-ID datasets caused by differences in, for example, lighting conditions, background and viewpoint (see Fig.~\ref{fig:reid_examples}), directly applying a re-ID model trained on a source dataset to an unseen target dataset will typically lead to large performance drops~\cite{zhong2019invariance,zhong2018gen,zhong2019camstyle,liu2019adaptive}. This suggests that the learned re-ID features severely overfit the source domain data and hence are not domain-generalisable. A domain-generalisable re-ID model has great values for real-world large-scale deployment. This is because such a model can work in any unseen scenarios, without the need to go through the tedious processes of data collection, annotation, and model updating/fine-tuning.

In this paper, we address both problems by designing novel CNN architectures. First, we argue that discriminative re-ID features need to be of \emph{omni-scale}, defined as the combination of variable homogeneous scales and heterogeneous scales, each of which is composed of a mixture of multiple scales. The need for omni-scale features is evident from Fig.~\ref{fig:reid_examples}. Specifically, to match people and distinguish them from distractors that cause false matches, the features corresponding to small local regions (e.g., shoes and glasses) and global whole body regions are equally important. For instance, given the query image in Fig.~\ref{fig:reid_examples}(a, left), looking at the global-scale features (e.g., young man, a white T-shirt + grey shorts combo) could narrow down the search to the true match (middle) and a distractor (right). Now the local-scale features come into play---the shoe region explains away the fact that the person on the right is a distractor (trainers vs.~sandals). However, for more challenging cases, even the features of variable homogeneous scales are not enough; more complicated and richer features that span multiple scales are required. For instance, to eliminate the distractor in Fig.~\ref{fig:reid_examples}(b, right), one needs the features that represent a white T-shirt with a specific logo in the front. Note that the logo is not distinctive on its own---without the white T-shirt as context, it can be confused with many other patterns. Moreover, the white T-shirt is likely everywhere in summer, e.g., Fig.~\ref{fig:reid_examples}(a). It is however the unique combination, captured by heterogeneous features spanning both small (logo size) and medium (upper body size) scales, that makes the features most effective.

\begin{figure}[t]
\centering
\includegraphics[width=0.8\columnwidth]{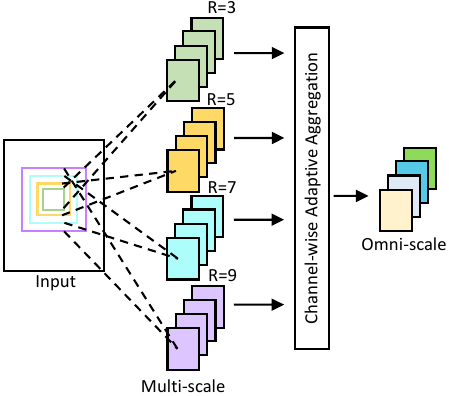}
\caption{A schematic of OSNet building block. R: Receptive field size.}
\label{fig:motivation}
\end{figure}

We therefore propose \emph{omni-scale network} (OSNet), a novel CNN architecture designed specifically for omni-scale feature learning. The underpinning building block of OSNet consists of multiple convolutional streams with different receptive field sizes\footnote{In this paper, `scale' and `receptive field' are used interchangeably.} (see Fig.~\ref{fig:motivation}). The feature scale that each stream focuses on is determined by \emph{exponent}, a new dimension factor that linearly increases across streams to ensure that various scales can be captured in each individual block. Critically, the resulting multi-scale feature maps are dynamically fused by channel-wise weights generated by a unified aggregation gate (AG). The AG is a mini-network sharing parameters across all streams with a number of desirable properties for effective model training. Since the AG are trainable, the generated channel-wise weights are input-dependent, realising a dynamic scale fusion. This novel AG design is crucial for learning omni-scale feature representations: conditioning on a specific input image, the gate can focus on a single scale by assigning a dominant weight to a particular stream or scale; alternatively, it can select and mix jointly to produce features with heterogeneous scales.

Another key characteristic of OSNet is \emph{lightweight}. A lightweight re-ID model has a couple of benefits: (1) Re-ID datasets are often of moderate size due to difficulties in collecting cross-camera matched person images. A lightweight network with a small number of parameters is thus less prone to overfitting; (2) In large-scale surveillance applications (e.g., city-wide surveillance with thousands of cameras), the most practical way for re-ID is to perform feature extraction at the camera end and send the extracted features to the central server rather than the raw videos. For on-device processing, small re-ID networks are clearly preferred. To this end, in our building block we factorise standard convolutions with pointwise and depthwise convolutions~\cite{howard2017mobilenets,sandler2018mobilenetv2}, making OSNet not only discriminative in feature learning but also efficient in implementation and deployment.

To address the second problem caused by domain gaps across different re-ID datasets, we notice that these gaps are typically reflected by different image styles, such as brightness, colour temperatures and view angles (see Fig.~\ref{fig:reid_examples}). These style variations are caused by differences in both lighting condition and camera characteristics/setup in different camera networks. Existing works address this problem using unsupervised domain adaptation (UDA) methods~\cite{zhong2019invariance,zhong2018gen,zhong2019camstyle,liu2019adaptive}. These require unlabelled target domain data to be available for model adaptation. In contrast, we treat this as a \emph{more general} domain generalisation (DG) problem~\cite{zhou2021domain} \emph{without using any target domain data}. By eliminating the tedious processes of data collection and model updating given a new target domain, our approach enables a re-ID model trained using source datasets to be applied out-of-the-box for any \emph{unseen} target dataset.

Concretely, our solution to domain-generalisable feature learning is to introduce instance normalisation (IN)~\cite{ulyanov2016instance} to our OSNet architecture. Unlike batch normalisation (BN)~\cite{ioffe2015batch} based on mini-batch level statistics, IN calibrates a sample using inner statistics, thus eliminating instance-specific contrast and style that are largely affected by domain-specific environments~\cite{ulyanov2017improved,dumoulin2016learned,AdaIN}. In this way, IN can naturally address the fundamental style discrepancy problem in cross-domain person re-ID. However, IN has never been exploited for solving such a cross-domain issue in re-ID. It has been noticed that where and how many IN layers to have in a CNN are critical for DG~\cite{pan2018ibn,nam2018BIN}, but there is no clear guidance on how to design the network architecture. We therefore propose to \emph{learn} the optimal model configuration directly from data via differentiable architecture search. More specifically, we design a novel search space, which contains candidate building blocks with different IN configurations. As the discrete selection variables disable search differentiation, we further leverage the Gumbel-Softmax~\cite{gumbelSoftmax,concreteDistribution} to create continuous representations for candidate selection, allowing end-to-end optimisation via gradient descent.

The contributions can be summarised as follows.
\textbf{(1)} We introduce, for the first time, the concept of omni-scale feature learning for discriminative person re-ID. This leads to OSNet, a novel CNN architecture capable of simultaneously learning \emph{homogeneous}- and \emph{heterogeneous}-scale features. By using factorised convolutions, OSNet is lightweight with only 2.2 million parameters---more than one order of magnitude smaller than the common ResNet50-based re-ID models.
\textbf{(2)} To improve domain generalisation cross datasets we incorporate instance normalisation (IN) into the OSNet design through differentiable architecture search, which we call OSNet-AIN. To our knowledge, this is the first work that explores both IN and neural architecture search for cross-domain re-ID.
\textbf{(3)} We evaluate OSNet by conducting extensive experiments on seven person re-ID datasets. In the same-domain setting, OSNet achieves state-of-the-art performance, outperforming many far larger re-ID models, often by a clear margin. Importantly, in the cross-domain setting, OSNet-AIN exhibits a remarkable generalisation ability: it beats most recent unsupervised domain adaptation (UDA) methods on unseen target domains while maintaining strong source domain performance, \emph{requiring neither the target domain data nor per-domain training}.
Our code and models are publicly available to facilitate future research in re-ID.\footnote{\texttt{https://github.com/KaiyangZhou/deep-person-reid}.}

\section{Related Work} \label{sec:relatedwork}

\keypoint{CNN Architectures for Person Re-ID}
Most existing deep re-ID methods~\cite{sun2018beyond,chang2018multi,li2018harmonious,si2018dual,hou2019interaction} adopt CNN architectures that are originally designed for generic object classification problems, especially those ImageNet-winning models~\cite{he2016deep,szegedy2015going,huang2017densely}. These architectures are intrinsically limited for instance recognition in re-ID. Modifications are thus made to tackle problems specific to re-ID, such as misalignment~\cite{wei2017glad,fu2019horizontal} and pose variations~\cite{hou2019interaction,shen2018end}. As persons usually stand upright, \cite{sun2018beyond,wang2018learning,fu2019horizontal} partition feature maps horizontally and inject parallel supervision signals to each stripe in order to enhance the learning of part-level features. Attention mechanisms are designed in~\cite{si2018dual,song2018mask,li2018harmonious} to focus feature learning on the foreground image regions. In~\cite{zhao2017spindle,su2017pose,xu2018attention,suh2018part,tian2018eliminating,zhang2019densely}, body part specific CNNs are learned by means of off-the-shelf pose detectors. In~\cite{li2017person,li2017learning,zhao2017deeply}, CNNs are branched to learn representations from global and local image regions. Since low-level visual cues such as colour are relevant for re-ID, \cite{yu2017devil,chang2018multi,liu2017hydraplus,wang2018resource} combine multi-level features extracted at different CNN layers. However, none of the existing re-ID networks can learn multi-scale features \emph{explicitly} at each CNN layer as in our OSNet. Unlike OSNet, they typically rely on external pose models and/or hand-pick some layers for multi-scale learning. Moreover, the ability to learn heterogeneous-scale features representing the mixture of different scales is also missing.

\keypoint{Multi-Scale and Multi-Stream CNNs}
As far as we know, the concept of omni-scale deep feature learning has never been introduced before. Nonetheless, the importance of multi-scale feature learning has been recognised recently, and the multi-stream building block design has also been adopted in re-ID~\cite{chen2017person}. Compared to a small number of re-ID networks that have multi-stream building blocks~\cite{chang2018multi,qian2017multi}, OSNet is significantly different. Specifically, the layer design in~\cite{chang2018multi} is based on ResNeXt~\cite{xie2017aggregated}, where each stream learns features at the same scale, while the streams in each OSNet block cover different scales. The network in~\cite{qian2017multi} is built on Inception~\cite{szegedy2015going,szegedy2016rethinking}, where the multi-streams were originally designed for low computational cost with a hand-crafted mixture of convolutional and pooling layers. In contrast, our building block uses a scale-controlling factor to diversify the spatial scales. Moreover, \cite{qian2017multi} fuses multi-stream features with learnable but fixed-once-learned stream-wise weights only at the final block. Whereas we fuse multi-scale features within each building block using dynamic (input-dependent) channel-wise weights to learn combinations of multi-scale patterns. Therefore, only our OSNet is capable of learning omni-scale features with each feature channel potentially capturing more discriminative features of either a single scale or a weighted mixture of multiple scales. Our experiments (in Sec.~\ref{subsec:sameDomainReID}) show that OSNet significantly outperforms the models in~\cite{chang2018multi,qian2017multi,chen2017person}.

\keypoint{Lightweight Network Design}
With embedded AI becoming topical, lightweight CNN design has attracted increasing attention. SqueezeNet~\cite{iandola2016squeezenet} compresses feature dimensions using $1\! \times\! 1$ convolutions. IGCNet~\cite{zhang2017interleaved}, ResNeXt~\cite{xie2017aggregated} and CondenseNet~\cite{huang2018condense} leverage group convolutions. Xception~\cite{chollet2017xception} and the MobileNet series~\cite{howard2017mobilenets,sandler2018mobilenetv2} are based on depthwise separable convolutions. Dense $1\! \times\! 1$ convolutions are grouped with channel shuffling in ShuffleNet~\cite{zhang2018shufflenet}. In terms of lightweight design, our OSNet is similar to MobileNet---both use factorised convolutions---but with a modification in the ordering that empirically works better for omni-scale feature learning (see Sec.~\ref{sec:depthwise_sep} for the details).

\keypoint{Domain Generalisation}
Cross-dataset generalisation has been studied in re-ID~\cite{song2019DGreid}, but no specific designs have ever been introduced to make re-ID models more \emph{intrinsically} generalisable. Recently, unsupervised domain adaptation (UDA) methods~\cite{zhong2019invariance,zhong2018gen,zhong2019camstyle,liu2019adaptive,zhou2020domain} have been extensively studied to adapt a re-ID model from source to target domain. However, UDA methods have to use unlabelled target domain data, so data collection and (per-domain) model update are still required. In contrast, without these steps, our OSNet-AIN is much more efficient in practice. Beyond re-ID, the problem of domain generalisation (DG) has been investigated in deep learning~\cite{li2017deeper,cvpr19JiGen,li2019episodic,zhou2020deep,zhou2020learning,zhou2021mixstyle} (see~\cite{zhou2021domain} for a comprehensive survey in this topic). However, most existing DG methods~\cite{cvpr19JiGen,li2019episodic,balaji2018metareg} assume that the source and target domains have the same label space, which apparently conflicts with the disjoint label space case in re-ID. Some recent few-shot meta-learning approaches are also re-purposed for DG~\cite{vinyals2016matching}. However, they assume a fixed number of classes for the target domain and are trained specifically for that number using source data. Therefore, they cannot be directly applied to re-ID, where the target domain has a different and variable number of identity classes.

Our DG-oriented re-ID solution is based on instance normalisation (IN) layers~\cite{ulyanov2016instance}. IN's ability to  eliminate instance-specific style discrepancy has been investigated for the style transfer task~\cite{ulyanov2017improved,dumoulin2016learned,AdaIN}. Recently, several works have attempted to integrate CNNs with IN layers to improve model generalisation. \cite{nam2018BIN} tackle multi-domain learning by fusing BN and IN with a convex weight. In~\cite{pan2018ibn}, an architecture called IBN-Net is engineered by inserting IN to shallow CNN layers for cross-domain semantic segmentation. The empirical study in~\cite{pan2018ibn} suggests that appearance variations mainly lie in shallow CNN layers, and therefore, inserting IN to shallow layers should be more effective. However, there is no clear definition of what `shallow' layers are in deep neural networks. Moreover, person re-ID is an instance recognition problem, for which the empirical rule derived from the semantic segmentation task might not work. In this paper, instead of hand-picking layers for inserting IN, we propose to use neural architecture search to optimally explore the capability of IN for improving DG.

\keypoint{Neural Architecture Search}
Neural architecture search (NAS) aims to automate the process of network architecture engineering. Early NAS methods are typically based on either reinforcement learning (RL)~\cite{NAS_RL,NASNet} or evolutionary algorithm (EA)~\cite{real2017large,real2019regularized}, where hundreds and thousands of models need to be trained from scratch and evaluated on a separate validation set to provide the supervision signal (reward for RL and fitness score for EA). This is computationally extremely expensive, requiring hundreds or even thousands of GPU days to complete the search. The follow-up research is mainly focused on accelerating the search by, for example, weight sharing~\cite{NAS_oneShot,ENAS}. Recently, there has been a growing interest in modelling NAS with directed acyclic graph (DAG) and using continuous representations for end-to-end optimisation. DARTS~\cite{Darts} uses softmax to relax the discrete one-hot actions over search space. Similarly, SNAS~\cite{SNAS} and GDAS~\cite{dong2019search} utilise a discrete gradient estimator~\cite{gumbelSoftmax,concreteDistribution} to overcome the non-differentiable nature in categorical variables. To address the high GPU memory problem in differentiable NAS, ProxylessNAS~\cite{cai2018proxylessnas} forces gradients to propagate through only one of the candidate paths. Different from these NAS methods, we do not search architecture from scratch. Instead, we base the architecture on OSNet and leverage NAS to find the best way to combine OSNet with IN.

An \textbf{earlier and preliminary version} of this work was published in ICCV'19~\cite{zhou2019osnet}. Compared with~\cite{zhou2019osnet}, which only focuses on discriminative feature learning for same-domain re-ID, this paper brings up the problem of generalisable feature learning for cross-domain re-ID, which has been largely overlooked in existing re-ID research. To that end, this work proposes to combine OSNet with IN by automatically searching for the best model configuration directly from data. Extensive experiments in the cross-domain re-ID setting, together with a comprehensive comparison with state-of-the-art cross-domain re-ID methods, demonstrate that our OSNet achieves strong performance on unseen target datasets even without 1) using any target domain data and 2) undesirable per-domain model adaptation steps.

\section{Omni-Scale Network for Person Re-ID} \label{sec:method}
In this section, we detail the design of our omni-scale network (OSNet), which is aimed at learning omni-scale feature representations for person re-ID. We first discuss depthwise separable convolutions, which are used to make OSNet lightweight. Then, we introduce our novel omni-scale residual block for learning discriminative re-ID features. Finally, to enhance generalisation in unseen datasets, we extend OSNet by adding instance normalisation (IN) layers and further present a differentiable architecture search mechanism to automatically infer the optimal IN configuration.

\subsection{Depthwise Separable Convolutions} \label{sec:depthwise_sep}
\begin{figure}[t]
\centering
\includegraphics[width=0.6\columnwidth]{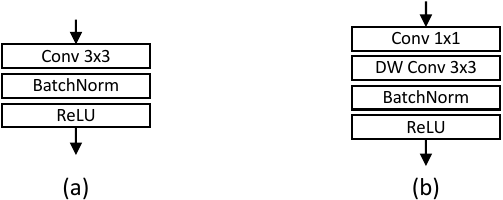}
\caption{(a) Standard and (b) Lite $3\! \times\! 3$ convolution. DW: Depth-Wise.}
\label{fig:liteconv}
\end{figure}

For lightweight network design, we adopt the depthwise separable convolution~\cite{howard2017mobilenets,chollet2017xception}. The basic idea is to divide a convolution layer, $\mathrm{ReLU} (\bm{w} * \bm{x})$, with kernel $\bm{w} \in \mathbb{R}^{k \times k \times c \times c'}$, into two separate layers, $\mathrm{ReLU} ((\bm{v} \circ \bm{u}) * \bm{x})$, with depthwise kernel $\bm{u} \in \mathbb{R}^{k \times k \times 1 \times c'}$ and pointwise kernel $\bm{v} \in \mathbb{R}^{1 \times 1 \times c \times c'}$, where $*$ denotes convolution, $k$ the kernel size, $c$ the input channel width and $c'$ the output channel width. Given an input tensor $\bm{x} \in \mathbb{R}^{h \times w \times c}$ of height $h$ and width $w$, the computational cost is reduced from $h \cdot w \cdot k^2 \cdot c \cdot c'$ to $h \cdot w \cdot (k^2 + c) \cdot c'$, and the parameter size from $k^2 \cdot c \cdot c'$ to $(k^2 + c) \cdot c'$. In our implementation, we find that $\mathrm{ReLU} ((\bm{u} \circ \bm{v}) * \bm{x})$ (pointwise $\to$ depthwise instead of depthwise $\to$ pointwise) turns out to be more effective for omni-scale feature learning.\footnote{The subtle difference between these two orders is when the channel width is increased: pointwise $\to$ depthwise increases the channel width before spatial aggregation.} We call such layer \emph{Lite $3\! \times \!3$} hereafter. The design is depicted in Fig.~\ref{fig:liteconv}.

\subsection{Omni-Scale Residual Block}
\begin{figure}[t]
\centering
\includegraphics[width=\columnwidth]{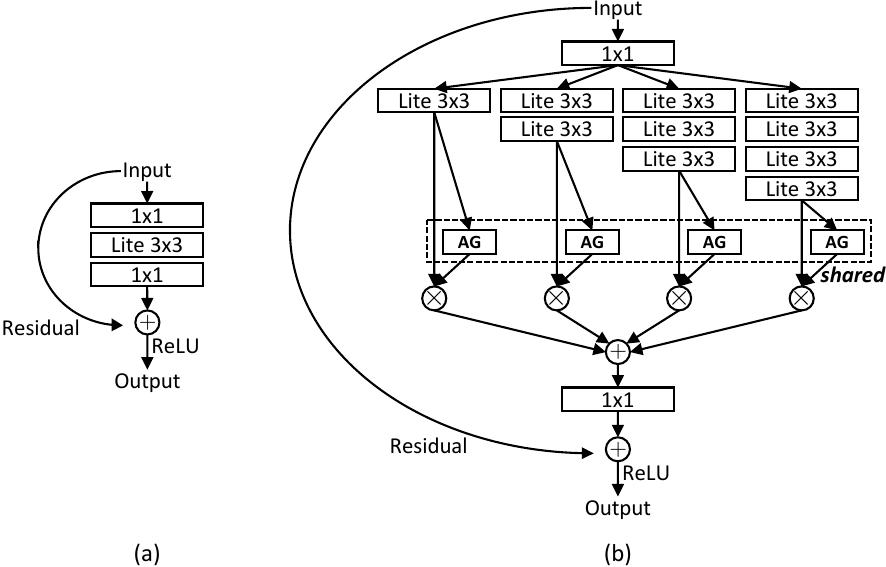}
\caption{
(a) Baseline bottleneck. (b) OSNet bottleneck. AG: Aggregation Gate. The first/last $1\! \times\! 1$ layers used to reduce/restore feature dimension.
}
\label{fig:bottleneck}
\end{figure}

The building block in OSNet is based on the residual bottleneck~\cite{he2016deep}, but equipped with the Lite $3\! \times \!3$ layer (Fig.~\ref{fig:bottleneck}). Given an input $\bm{x}$, a residual bottleneck aims to learn a residual $\tilde{\bm{x}}$ via a mapping function $F$, i.e.
\begin{equation} \label{eq:base_residual}
\bm{y} = \bm{x} + \tilde{\bm{x}}, \quad \text{with} \quad \tilde{\bm{x}} = F(\bm{x}), 
\end{equation}
where $F$ denotes a Lite $3\! \times\! 3$ convolution layer that learns \emph{single-scale} features (i.e.~receptive field = $3\! \times\! 3$). Note that here the $1\! \times\! 1$ convolution layers are ignored in notation as they are used to manipulate feature channels and do not contribute to the aggregation of spatial information~\cite{he2016deep,xie2017aggregated}.

\keypoint{Multi-Scale Feature Learning}
To achieve multi-scale feature learning, we extend the residual function $F$ by introducing a new dimension, \emph{exponent} $t$, to represent the feature scale. For $F^t$, with $t > 1$, we stack $t$ Lite $3\! \times\! 3$ layers, resulting in a receptive field of size $(2t+1)\! \times\! (2t+1)$. Then, the residual to be learned, $\tilde{\bm{x}}$, is the sum of incremental scales of representations up to $T$:
\begin{equation} \label{eq:mult_stream_residual}
\tilde{\bm{x}} = \sum_{t=1}^T F^t(\bm{x}), \quad \text{} \quad T \geqslant 1.
\end{equation}

When $T\! =\! 1$, Eq.~\eqref{eq:mult_stream_residual} reduces to Eq.~\eqref{eq:base_residual}, i.e. the baseline single-scale bottleneck as shown in Fig.~\ref{fig:bottleneck}(a). Considering the computational cost, we use $T\! =\! 4$ in this paper where the largest receptive field is $9\! \times\! 9$. This is depicted in Fig.~\ref{fig:bottleneck}(b).

\keypoint{Dynamic and Unified Aggregation Gate}
So far, each individual stream gives us features of only one specific scale, i.e. \emph{scale homogeneous}. To learn effective omni-scale features, we propose to combine the outputs of different streams in a \emph{dynamic} way, i.e.~different weights are assigned to different scales according to the input image, rather than being fixed and identical for all the data after training. More specifically, the dynamic scale fusion is achieved by a novel aggregation gate (AG), which is essentially a \emph{learnable neural network}.

Let $\bm{x}^t$ denote $F^t (\bm{x})$, the omni-scale residual $\tilde{\bm{x}}$ is then formulated by
\begin{equation} \label{eq:ag}
\tilde{\bm{x}} = \sum_{t=1}^T G(\bm{x}^t) \odot \bm{x}^t, \quad \text{with} \quad \bm{x}^t \triangleq F^t(\bm{x}),
\end{equation}
where $G (\bm{x}^t)$ is a data-conditioned vector with length spanning the entire channel dimension of input $\bm{x}^t$, and $\odot$ denotes the Hadamard product. $G$ is implemented as a mini-network composed of a non-parametric global average pooling layer~\cite{lin2013network} and a multi-layer perceptron (MLP) with one ReLU-activated hidden layer, followed by the sigmoid activation. To reduce parameter overhead, we follow~\cite{woo2018cbam,hu2018senet} to reduce the MLP's hidden dimension with a reduction ratio, which is set to 16.

In our design, the AG is \emph{shared} across all the feature streams in the same omni-scale residual block (dashed box in Fig.~\ref{fig:bottleneck}(b)). In spirit, this is similar to the parameter sharing of convolution filters in CNNs, resulting in a number of advantages. First, the number of parameters is independent of the number of streams $T$, thus the model becomes more scalable. Second, unifying AG has a nice property when performing gradient backpropagation. Concretely, suppose the network is supervised by a differentiable loss function $\mathcal{L}$ and the gradient $\frac{\partial \mathcal{L}}{\partial \tilde{\bm{x}}}$ can be computed. The gradient w.r.t $G$, based on Eq.~\eqref{eq:ag}, is
\begin{equation} \label{eq:grad_ag}
\frac{\partial \mathcal{L}}{\partial G} = \frac{\partial \mathcal{L}}{\partial \tilde{\bm{x}}} \frac{\partial \tilde{\bm{x}}}{\partial G}
= \frac{\partial \mathcal{L}}{\partial \tilde{\bm{x}}} ( \sum_{t=1}^T \bm{x}^t ).
\end{equation}
It is clear that the second term in Eq.~\eqref{eq:grad_ag} indicates that supervision signals from all streams are gathered together to guide the learning of $G$. This desirable property disappears when each stream has its own gate.

We further stress two design considerations. First, in contrast to using a single-scalar gate function that provides a coarse scale fusion, we use \emph{channel-wise vector} gating, i.e. AG's output $G (\bm{x}^t)$ is a vector rather a scalar for the $t$-th stream. This design results in a more fine-grained fusion that tunes each feature channel. Second, the weights are \emph{dynamically} computed by conditioning on the input data. This is crucial for re-ID as the training and test data describe disjoint identity populations; input adaptive feature-scale fusion is hence more effective and scalable.

\subsection{Inserting Instance Normalisation Layers} \label{subsec:searchOSNetIN}
Different from batch normalisation (BN)~\cite{ioffe2015batch}, which normalises each sample using statistics computed over a mini-batch, IN performs normalisation on each sample using its own mean and standard deviation~\cite{ulyanov2017improved}.
As such, IN allows the instance-specific style information to be effectively removed. Therefore inserting IN layers into a re-ID CNN has the potential of eliminating image style differences caused by distinct environments, lighting conditions, camera setups, etc. in each dataset. However, it is unclear how to integrate a re-ID CNN with IN to maximise the gain, e.g., which layers to insert? Inside or outside a residual block?

\begin{figure}[t]
    \centering
    \includegraphics[width=.98\columnwidth]{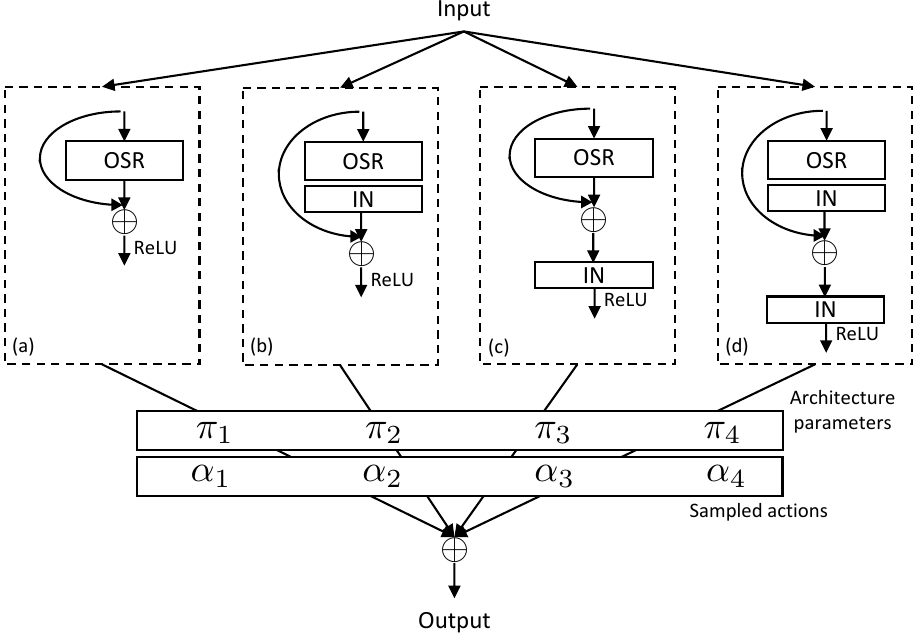}
    \caption{Our architecture search space consists of four different omni-scale residual (OSR) blocks each with a learnable parameter $\pi$. During a forward pass, the selected candidate is determined by sampling discrete actions (one-hot) from a categorical distribution parameterised by the architecture parameters. To make the computational graph differentiable, we relax the discrete variables to continuous representations using the Gumbel-Softmax~\cite{gumbelSoftmax,concreteDistribution}. IN: Instance Normalisation~\cite{ulyanov2016instance}.}
    \label{fig:search_space}
\end{figure}

\keypoint{Architecture Search Space}
We propose to learn the optimal way of integrating OSNet with IN by neural architecture search (NAS). To that end, we define a novel search space $\Omega$ consisting of candidate omni-scale (OS) blocks in different IN-incorporating designs. Specifically, besides the standard OS block (Fig.~\ref{fig:bottleneck}(b)), we design three other variants (see Fig.~\ref{fig:search_space} for an illustration of our search space). Following~\cite{he2016identity}, we keep the residual learning module unchanged, i.e.~only adding IN after the residual. For clarity, we refer to Fig.~\ref{fig:search_space}(b-d) as OS+IN$_{\mathrm{in}}$ block, OS+IN$_{\mathrm{out}}$ block, and OS+IN$_{\mathrm{in-out}}$ block, respectively.

\keypoint{Formulation}
In our NAS formulation, the output $\bm{y}$ of each OS layer is obtained as a weighted sum of operations in $\Omega$,
\begin{equation} \label{eq:search_space_weighted_sum}
\bm{y} = \sum_{\omega \in \Omega} \alpha_\omega \omega(\bm{x}),
\end{equation}
where $\bm{\alpha} = [\alpha_\omega]_{\omega \in \Omega}$ is a $|\Omega|$-dimensional one-hot vector, with the activated element ``1'' corresponding to the IN design selection.

The objective is to minimise the following expectation through jointly optimising the model architecture $\bm{\alpha}$ and the parameters $\bm{\theta}$ as,
\begin{equation} \label{eq:raw_nas_obj}
\mathbb{E} 
[\mathcal{L}(\bm{x}, \bm{\theta}, \bm{\alpha})].
\end{equation}

For an OSNet with $m$ blocks, the search space contains a total of $4^m$ different architecture design choices. A key challenge lies in the optimisation of the discrete selection that is non-differentiable due to its discontinuous nature. This disables the adoption of strong gradient-based architecture search optimisation.

\keypoint{Relaxation and Reparameterisation Trick}
To solve the non-differentiable problem, we develop a continuous relaxation and reparameterisation strategy. This is achieved by first treating $\bm{\alpha}$ as a continuous $\ell_1$-normalised random variable sampled from a probability distribution $P_{\bm{\pi}}$ parameterised by $\bm{\pi}$ (i.e.~the target architecture parameters) as
\begin{equation} \label{eq:softmax_prob}
P(\alpha_\omega = 1) = \frac{\exp (\pi_\omega)}{\sum_{\omega' \in \Omega} \exp(\pi_{\omega'})},
\end{equation}
and then reparameterising this sampling process by the Gumbel-Softmax~\cite{gumbelSoftmax,concreteDistribution} defined as
\begin{equation} \label{eq:gumbel_softmax}
\alpha_\omega = f_{\bm{\pi}}(z_\omega) = \frac{\exp\big((\log \pi_\omega + z_\omega) / \lambda\big)}{\sum_{\omega' \in \Omega} \exp\big((\log \pi_{\omega'} + z_{\omega'}) / \lambda\big)},
\end{equation}
where $\lambda$ is the softmax temperature and $z_\omega \sim \mathrm{Gumbel}(0, 1)$ a Gumbel distribution. Concretely, $z_\omega$ is obtained by the following transformation of the uniform distribution: $z_\omega = - \log (- \log (u_\omega))$, where $u_\omega \sim \mathrm{Uniform}(0, 1)$.

The objective function is reformulated as:
\begin{equation} \label{eq:reparam_nas_obj}
\mathbb{E}_{\bm{z} \sim P(\bm{z})}[\mathcal{L}(\bm{x}, \bm{\theta}, f_{\bm{\pi}}(\bm{z}))],
\end{equation}
which is fully differentiable w.r.t.~both $\bm{\theta}$ and $\bm{\pi}$. The gradients can be approximated by Monte Carlo sampling~\cite{mohamed2019monte},
\begin{align}\small
& \triangledown_{\bm{\theta}} \mathbb{E}_{\bm{z} \sim P(\bm{z})}[\mathcal{L}(\bm{x}, \bm{\theta}, f_{\bm{\pi}}(\bm{z}))] \\
= \; & \mathbb{E}_{\bm{z} \sim P(\bm{z})}[\triangledown_{\bm{\theta}} \mathcal{L}(\bm{x}, \bm{\theta}, f_{\bm{\pi}}(\bm{z}))] \\
\simeq \; & \frac{1}{S} \sum_{s=1}^S \triangledown_{\bm{\theta}} \mathcal{L}(\bm{x}, \bm{\theta}, f_{\bm{\pi}}(\bm{z}^s))], \label{eq:mc_grad_theta}
\end{align}
where $S$ denotes the number of sampling steps, and similarly,
\begin{align}
& \triangledown_{\bm{\pi}} \mathbb{E}_{\bm{z} \sim P(\bm{z})}[\mathcal{L}(\bm{x}, \bm{\theta}, f_{\bm{\pi}}(\bm{z}))] \\
= \; & \mathbb{E}_{\bm{z} \sim P(\bm{z})}[\triangledown_{\bm{\pi}} \mathcal{L}(\bm{x}, \bm{\theta}, f_{\bm{\pi}}(\bm{z}))] \\
\simeq \; & \frac{1}{S} \sum_{s=1}^S \triangledown_{\bm{\pi}} \mathcal{L}(\bm{x}, \bm{\theta}, f_{\bm{\pi}}(\bm{z}^s))]. \label{eq:mc_grad_pi}
\end{align}

In doing so, we transfer the dependency on $\bm{\pi}$ from $P$ to $f$. Importantly, as proved in \cite{concreteDistribution}, when $\lambda \rightarrow 0$, the relaxed softmax formulation (Eq.~\eqref{eq:gumbel_softmax}) approaches the discrete $\mathrm{argmax}$ computation, i.e.~an unbiased gradient estimator.

\keypoint{Search Outcome}
At the end of search, we derive a compact network architecture by selecting for each layer the OS block with the largest $\pi$, i.e.~$\omega^* = \arg\max_{\omega \in \Omega} \pi_\omega$.

\begin{table}[t]
\setlength{\tabcolsep}{10pt}
\renewcommand{\arraystretch}{\tableCellHeight}
\renewcommand{\arraystretch}{1.2}
\newcommand{\numCols}{2}
\caption{Omni-scale network architecture for person re-ID. Input image size is $256\! \times\! 128$. AIN: Automatic search + Instance Normalisation.}
\label{tab:netarch}
\centering
\footnotesize
\begin{tabular}{c | c | c | c}
\toprule
stage & output & OSNet & OSNet-AIN \\ \midrule
conv1 & 128$\! \times\! $64, 64 & \multicolumn{\numCols}{c}{7$\! \times\! $7 conv, stride 2} \\
pool & 64$\! \times\! $32, 64 & \multicolumn{\numCols}{c}{3$\! \times\! $3 max pool, stride 2} \\ \midrule

\multirow{2}{*}{conv2} & 64$\! \times\! $32, 256 & OS block & OS+IN$_{\mathrm{in}}$ block \\
 & 64$\! \times\! $32, 256 & OS block & OS+IN$_{\mathrm{in}}$ block \\ \midrule

\multirow{2}{*}{transition} & 64$\! \times\! $32, 256 & \multicolumn{\numCols}{c}{1$\! \times\! $1 conv} \\
 & 32$\! \times\! $16, 256 & \multicolumn{\numCols}{c}{2$\! \times\! $2 average pool, stride 2} \\ \midrule

\multirow{2}{*}{conv3} & 32$\! \times\! $16, 384 & OS block & OS block \\
 & 32$\! \times\! $16, 384 & OS block & OS+IN$_{\mathrm{in}}$ block \\ \midrule

\multirow{2}{*}{transition} & 32$\! \times\! $16, 384 & \multicolumn{\numCols}{c}{1$\! \times\! $1 conv} \\
 & 16$\! \times\! $8, 384 & \multicolumn{\numCols}{c}{2$\! \times\! $2 average pool, stride 2} \\ \midrule

\multirow{2}{*}{conv4} & 16$\! \times\! $8, 512 & OS block & OS+IN$_{\mathrm{in}}$ block \\
 & 16$\! \times\! $8, 512 & OS block & OS block \\ \midrule

conv5 & 16$\! \times\! $8, 512 & \multicolumn{\numCols}{c}{1$\! \times\! $1 conv} \\ \midrule
gap & 1$\! \times\! $1, 512 & \multicolumn{\numCols}{c}{global average pool} \\ \midrule
fc & 1$\! \times\! $1, 512 & \multicolumn{\numCols}{c}{fc} \\ \midrule

\multicolumn{2}{c|}{\# params} & 2.2M & 2.2M \\
\midrule
\multicolumn{2}{c|}{Mult-Adds} & 978.9M & 978.9M \\
\bottomrule
\end{tabular}
\end{table}

\subsection{Network Architecture}
\keypoint{OSNet}
is constructed by stacking the proposed lightweight bottleneck (OS block) layer-by-layer. The detailed network architecture is shown in Table~\ref{tab:netarch}. For comparison, the same network architecture with normal convolutions has 6.9 million parameters and 3,384.9 million mult-add operations. This is $3\! \times$ larger than OSNet with the Lite $3\! \times\! 3$ convolution layer design. The OSNet architecture in Table~\ref{tab:netarch} can be easily scaled up or down in practice, to balance model size, computational cost and performance. To this end, we use a width multiplier\footnote{Width multiplier with magnitude smaller than 1 works on all layers in OSNet except the last fc layer whose feature dimension is fixed to 512.} and an image resolution multiplier, following~\cite{howard2017mobilenets,sandler2018mobilenetv2,zhang2018shufflenet}.

\keypoint{OSNet-AIN}
denotes the network architecture with automatically searched IN layers. In the experiment, we run the searching algorithm four times with different random seeds and select the one with the best cross-domain performance as our final model, which is a commonly adopted protocol in the NAS literature~\cite{Darts,SNAS,cai2018proxylessnas}. The found best architecture model is shown in the OSNet-AIN column in Table~\ref{tab:netarch}. The detailed experimental setup for NAS will be covered in Sec.~\ref{subsec:crossDomainReID}. As IN only introduces a small number of parameters, the complexity between OSNet-AIN and OSNet is similar.

\keypoint{Relation to Prior Architectures}
In terms of the multi-stream design, OSNet is related to Inception~\cite{szegedy2015going} and ResNeXt~\cite{xie2017aggregated}, but has several crucial differences. 1) First, the multi-stream design in OSNet strictly follows the scale-incremental principle dictated by the exponent variable (see Eq.~\eqref{eq:mult_stream_residual}). Such a design is more effective for covering a wide range of scales. In contrast, Inception was originally designed to have a low computational cost by sharing computations with multiple streams. Therefore, its structure, which includes mixed operations of convolution and pooling, was hand-crafted. ResNeXt has multiple equal-scale streams, thus learning features at the same scale. 2) Second, Inception/ResNeXt aggregates features by concatenation/addition while OSNet uses the unified AG (Eq.~\eqref{eq:ag}) to facilitate the learning of heterogeneous-scale features. Critically, this means that the fusion in OSNet is dynamic and adaptive to each individual input image, which is more effective in dealing with disjoint label space in person re-ID. 3) Third, OSNet uses factorised convolutions and thus the building block and subsequently the whole network is lightweight. Therefore, the OSNet architecture is fundamentally different from Inception and ResNeXt in nature and design. While the AG borrows the design from SENet~\cite{hu2018senet}, they differ conceptually with separate purposes. SENet aims to re-calibrate feature channels by re-scaling the activation values for a single stream. Whereas, OSNet aims to selectively fuse multiple feature streams of different receptive field sizes for learning omni-scale features.

\section{Experiments} \label{sec:experiments}

\begin{table}[t]
\setlength{\tabcolsep}{10pt}
\renewcommand{\arraystretch}{\tableCellHeight}
\caption{Statistics of person re-ID datasets.}
\label{tab:reid_datasets_stats}
\centering
\footnotesize
\begin{tabular}{l | c c c}
\toprule
Dataset & \# IDs & \# images & \# cameras \\
\midrule
Market1501~\cite{zheng2015scalable} & 1,501 & 32,668 & 6 \\
CUHK03~\cite{li2014deepreid} & 1,467 & 28,192 & 2 \\
Duke~\cite{ristani2016performance,zheng2017unlabeled} & 1,812 & 36,411 & 8 \\
MSMT17~\cite{wei2018person} & 4,101 & 126,411 & 15 \\
VIPeR~\cite{gray2007evaluating} & 632 & 1,264 & 2 \\
GRID~\cite{loy2009multi} & 251 & 1,275 & 6 \\
CUHK01~\cite{CUHK01} & 971 & 3,882 & 2 \\
\bottomrule
\end{tabular}
\end{table}

\subsection{Same-Domain Person Re-Identification} \label{subsec:sameDomainReID}
We first evaluate OSNet in the conventional person re-ID setting where the model is trained and tested on the same dataset (domain).

\keypoint{Datasets and Settings}
Seven popular re-ID benchmarks are used, including Market1501~\cite{zheng2015scalable}, CUHK03~\cite{li2014deepreid}, DukeMTMC-reID (Duke)~\cite{ristani2016performance,zheng2017unlabeled}, MSMT17~\cite{wei2018person},\footnote{Throughout this paper, we use the v1 version for MSMT17.} VIPeR~\cite{gray2007evaluating}, GRID~\cite{loy2009multi} and CUHK01~\cite{CUHK01}. The overall dataset statistics are detailed in Table~\ref{tab:reid_datasets_stats}. The first four are typically considered as \emph{big} re-ID datasets---even though their sizes are fairly moderate (around 30k training images for the largest dataset MSMT17). The rest three datasets are generally too \emph{small} to train deep models without proper pre-training~\cite{li2017person,zhao2017spindle}. For CUHK03, we use the 767/700 split~\cite{zhong2017rerank} with the detected images. For VIPeR, GRID and CUHK01 (485/486 split~\cite{liao2015person}), we follow~\cite{liu2017hydraplus,wei2017glad,zhao2017spindle,li2017person,zhao2017deeply} to perform pre-training on large re-ID datasets and then fine-tune the model on the target dataset, where the results are averaged over 10 random splits. For evaluation metrics, we use cumulative matching characteristics (CMC) rank accuracy and mean average precision (mAP). The performance is reported in percentage.

\begin{table*}[h]
\setlength{\tabcolsep}{6pt}
\renewcommand{\arraystretch}{\tableCellHeight}
\centering
\footnotesize
\caption{Results on big re-ID datasets. It is noteworthy that OSNet surpasses most published methods by a clear margin
on all datasets with only 2.2 million parameters, far less than the best-performing ResNet-based methods. -: not reported. $\dag$: reproduced by us.
}
\label{tab:same_domain_big}
\begin{tabular}{l|c|c|c|cc|cc|cc|cc}
\toprule
\multirow{2}{*}{Method} & \multirow{2}{*}{Venue} & \multirow{2}{*}{Backbone}& \multirow{2}{*}{Params (M)} & \multicolumn{2}{c|}{Market1501} & \multicolumn{2}{c|}{CUHK03} & \multicolumn{2}{c|}{Duke} & \multicolumn{2}{c}{MSMT17} \\
                                        & & & & R1 & mAP & R1 & mAP & R1 & mAP & R1 & mAP \\
\midrule
\multicolumn{11}{c}{\it Trained from scratch} \\
\midrule
MobileNetV2$^{\dag}$~\cite{sandler2018mobilenetv2} & CVPR'18 & MobileNetV2 & 2.2 & 87.0 & 69.5 & {46.5} & {46.0} & 75.2 & 55.8 & {50.9} & {27.0} \\
BraidNet~\cite{wang2018person}    & CVPR'18 & BraidNet & - & 83.7 & 69.5 & - & - & 76.4 & 59.5 & - & - \\
HAN~\cite{li2018harmonious}   & CVPR'18 & Inception & 4.5 & {91.2} & {75.7} & 41.7 & 38.6 & {80.5} & {63.8} & - & - \\
Auto-ReID~\cite{autoreid}     & ICCV'19 & Auto & 13.1 & 90.7 & 74.6 & - & - & - & - & - & - \\
OSNet (\emph{ours}) & This work & OSNet & 2.2 & \textbf{93.6} & \textbf{81.0} & \textbf{57.1} & \textbf{54.2} & \textbf{84.7} & \textbf{68.6} & \textbf{71.0} & \textbf{43.3} \\ 
\midrule
\multicolumn{11}{c}{\it Pre-trained on ImageNet} \\
\midrule
SVDNet~\cite{sun2017svdnet}             & ICCV'17 & ResNet & $>$23.5 & 82.3 & 62.1 & 41.5 & 37.3 & 76.7 & 56.8 & - & - \\
PDC~\cite{su2017pose}                   & ICCV'17 & Inception & $>$6.8 & 84.1 & 63.4 & - & - & - & - & 58.0 & 29.7 \\
HAP2S~\cite{yu2018hard}                  & ECCV'18 & ResNet & $>$23.5 & 84.6 & 69.4 & - & - & 75.9 & 60.6 & - & - \\
DPFL~\cite{chen2017person}               & ICCVW'17 & Inception & $>$6.8 & 88.6 & 72.6 & 40.7 & 37.0 & 79.2 & 60.6 & - & - \\
DaRe~\cite{wang2018resource}          & CVPR'18 & DenseNet & $>$23.5 & 89.0 & 76.0 & 63.3 & 59.0 & 80.2 & 64.5 & - & - \\
PNGAN~\cite{qian2018pose}               & ECCV'18 & ResNet & $>$23.5 & 89.4 & 72.6 & - & - & 73.6 & 53.2 & - & - \\
GLAD~\cite{wei2017glad}                  & ACM MM'17 & Inception & $>$6.8 & 89.9 & 73.9 & - & - & - & - & 61.4 & 34.0 \\
KPM~\cite{shen2018end}                   & CVPR'18 & ResNet & $>$23.5 & 90.1 & 75.3 & - & - & 80.3 & 63.2 & - & - \\
MLFN~\cite{chang2018multi}               & CVPR'18 & ResNeXt & 32.5 & 90.0 & 74.3 & 52.8 & 47.8 & 81.0 & 62.8 & - & - \\
FDGAN~\cite{ge2018fd}                  & NeurIPS'18 & ResNet & $>$23.5 & 90.5 & 77.7 & - & - & 80.0 & 64.5 & - & - \\
DuATM~\cite{si2018dual}                 & CVPR'18 & DenseNet & $>$7.0 & 91.4 & 76.6 & - & - & 81.8 & 64.6 & - & - \\
Bilinear~\cite{suh2018part}             & ECCV'18 & Inception & $>$6.8 & 91.7 & 79.6 & - & - & 84.4 & 69.3 & - & - \\
G2G~\cite{shen2018deep}                  & CVPR'18 & ResNet & $>$23.5 & 92.7 & 82.5 & - & - & 80.7 & 66.4 & - & - \\
DeepCRF~\cite{chen2018group}          & CVPR'18 & ResNet & 26.1 & 93.5 & 81.6 & - & - & 84.9 & 69.5 & - & - \\
PCB~\cite{sun2018beyond}              & ECCV'18 & ResNet & 27.2 & 93.8 & 81.6 & 63.7 & 57.5 & 83.3 & 69.2 & 68.2 & 40.4 \\
SGGNN~\cite{shen2018person}            & ECCV'18 & ResNet & $>$23.5 & 92.3 & 82.8 & - & - & 81.1 & 68.2 & - & - \\
Mancs~\cite{wang2018mancs}             & ECCV'18 & ResNet & $>$25.1 & 93.1 & 82.3 & 65.5 & 60.5 & 84.9 & 71.8 & - & - \\
AANet~\cite{tay2019aanet}             & CVPR'19 & ResNet & $>$23.5 & 93.9 & 83.4 & - & - & {87.7} & {74.3} & - & - \\
CAMA~\cite{yang2019towards}            & CVPR'19 & ResNet & $>$23.5 & {94.7} & 84.5 & {66.6} & {64.2} & 85.8 & 72.9 & - & - \\
IANet~\cite{hou2019interaction}        & CVPR'19 & ResNet & $>$23.5 & 94.4 & 83.1 & - & - & 87.1 & 73.4 & 75.5 & 46.8 \\
DGNet~\cite{zheng2019joint}            & CVPR'19 & ResNet & $>$23.5 & \textbf{94.8} & {86.0} & 65.6 & 61.1 & 86.6 & {74.8} & {77.2} & {52.3} \\
Auto-ReID~\cite{autoreid}              & ICCV'19 & Auto & 13.1 & 94.5 & {85.1} & \textbf{73.3} & \textbf{69.3} & - & - & {78.2} & {52.5} \\
OSNet (\emph{ours}) & This work & OSNet & 2.2 & \textbf{94.8} & \textbf{86.7} & {72.3} & {67.8} & \textbf{88.7} & \textbf{76.6} & \textbf{79.1} & \textbf{55.1} \\ 
\bottomrule
\end{tabular}
\end{table*}

\keypoint{Implementation Details}
A classification layer (linear fc + softmax) is mounted on the top of OSNet. The training follows the standard classification paradigm where each person identity is regarded as a unique class. Similar to~\cite{li2018harmonious,chang2018multi}, the cross-entropy loss with label smoothing~\cite{szegedy2016rethinking} is used for supervision. For fair comparison against existing methods, we implement two versions of OSNet. One is trained from scratch while the other is fine-tuned from ImageNet pre-trained weights. Person matching is based on the cosine distance using 512-D feature vectors extracted from the last fc layer. The batch size and the weight decay are set to 64 and 5e-4 respectively. Images are resized to $256 \times 128$.

For training from scratch, SGD is used to optimise the network for 350 epochs. The learning rate starts from 0.065 and is decayed by 0.1 at the 150-th, the 225-th, and the 300-th epoch, respectively. Data augmentation includes random flip, random crop and random patch.\footnote{RandomPatch works by (1) constructing a patch pool that stores randomly extracted image patches and (2) pasting a random patch selected from the patch pool onto an input image at random position.} For fine-tuning, we train the network with AMSGrad~\cite{reddi2018on} and the initial learning rate of 0.0015 for 250 epochs. The learning rate is decayed using the cosine annealing strategy~\cite{cosineLR} (without restart). During the first 10 epochs, the ImageNet pre-trained base network is frozen. Only the randomly initialised classifier is open for training~\cite{geng2016deep}. Data augmentation includes random flip and random erasing~\cite{zhong2017random}.

\keypoint{Results on Big Re-ID Datasets}
From Table~\ref{tab:same_domain_big}, we have the following observations.
{\bf(1)} OSNet achieves the best overall performance, outperforming most recently published methods by a clear margin. It is evident that the performance on re-ID benchmarks, especially Market1501 and Duke, has been saturated lately. Therefore, the improvements obtained by OSNet are significant. Crucially, the improvements are achieved with \emph{much smaller model size}---most top-performing re-ID models are based on the ResNet50 backbone, which has more than 23.5 million parameters (except extra customised modules), whereas our OSNet has only 2.2 million parameters. Notably, OSNet is around $6\! \times$ smaller than the automatically searched model, Auto-ReID \cite{autoreid}, but obtains better performance on three out of four datasets. These results verify the effectiveness of omni-scale feature learning for re-ID, achieved by an extremely compact network. As OSNet is orthogonal to some methods such as the image generation-based DGNet~\cite{zheng2019joint}, they can be combined to potentially boost the re-ID performance in practice.
{\bf(2)} OSNet yields strong performance with or without ImageNet pre-training. Among the very few existing lightweight re-ID models that can be trained from scratch (Auto-ReID, HAN and BraidNet in the top group), OSNet exhibits more significant advantages. For instance, in terms of mAP, on Market1501, OSNet beats Auto-ReID, HAN and BraidNet by 6.4\%, 5.3\% and 11.5\%, respectively. Compared with MobileNetV2, which is a general-purpose lightweight CNN, OSNet achieves a large margin consistently at a similar model size. Overall, these results demonstrate the versatility of OSNet: it enables effective feature tuning from generic object categorisation tasks, and offers robustness against model overfitting when trained from scratch on datasets of moderate size.
{\bf(3)} Compared with re-ID models~\cite{li2018harmonious,su2017pose,chen2017person,wei2017glad,chang2018multi,suh2018part} also based on multi-scale/multi-stream architectures, namely Inception or ResNeXt, OSNet is clearly better. As discussed in Sec.~\ref{sec:method}, this is attributed to the unique ability of OSNet to learn heterogeneous-scale features by combining multiple homogeneous-scale features with the dynamic and unified AG.

\keypoint{Results on Small Re-ID Datasets}
Small re-ID datasets are more challenging for deep re-ID models because they have much less training images and classes than the big datasets. Table~\ref{tab:same_domain_small} compares OSNet with six state-of-the-art deep re-ID methods. On VIPeR, we observe that OSNet outperforms all alternatives by a significant margin (more than 11\%). GRID is even more challenging than VIPeR because it has only 250 training images of 125 identities. Moreover, it was captured by real (operational) analogue CCTV cameras installed in busy public spaces, presenting more observation noise. OSNet remains the best on GRID, marginally above JLML~\cite{li2017person}, which is the current state-of-the-art. On CUHK01, which has around 1,900 training images, OSNet significantly outperforms Spindle and JLML by 6.7\% and 16.8\%, respectively. Overall, the performance of OSNet on these small datasets is excellent, indicating its promising advantage in real-world applications \emph{without} large-scale training data.

\begin{table}[t]
\setlength{\tabcolsep}{5pt}
\renewcommand{\arraystretch}{\tableCellHeight}
\centering
\footnotesize
\caption{Comparison with deep methods on small re-ID datasets at rank-1.}
\label{tab:same_domain_small}
\begin{tabular}{l|c|c|c|c}
\toprule
Method & Backbone & VIPeR & GRID & CUHK01 \\
\midrule
MuDeep~\cite{qian2017multi} & Inception & 43.0 & - & - \\ 
DeepAlign~\cite{zhao2017deeply} & Inception & 48.7 & - & - \\ 
JLML~\cite{li2017person} & ResNet & 50.2 & 37.5 & 69.8 \\ 
Spindle~\cite{zhao2017spindle} & Inception & 53.8 & - & 79.9 \\ 
GLAD~\cite{wei2017glad} & Inception & 54.8 & - & - \\ 
HydraPlus-Net~\cite{liu2017hydraplus} & Inception & 56.6 & - & - \\ 
OSNet (\emph{ours}) & OSNet & \textbf{68.0} & \textbf{38.2} & \textbf{86.6} \\
\bottomrule
\end{tabular}
\end{table}

\begin{table}[t]
\setlength{\tabcolsep}{2pt}
\renewcommand{\arraystretch}{\tableCellHeight}
\centering
\footnotesize
\caption{Ablation study on omni-scale residual learning.}
\label{tab:ablation_omni_scale}
\begin{tabular}{c | l | c c | c c}
\toprule
\multirow{2}{*}{Model} & \multicolumn{1}{c|}{\multirow{2}{*}{Architecture}} & \multicolumn{2}{c|}{CUHK03} & \multicolumn{2}{c}{MSMT17} \\
 & & R1 & mAP & R1 & mAP \\
\midrule
\ablStdModel \label{ablStdModel:4str+uag} & $T=4$ + unified AG (\textbf{primary model}) & 57.1 & 54.2 & 71.0 & 43.3 \\
\ablStdModel \label{ablStdModel:4str(fullconv)+uag} & $T=4$ w/ full conv + unified AG & 59.1 & 56.2 & 70.5 & 43.2 \\
\ablStdModel \label{ablStdModel:4str(t=1)+uag} & $T=4$ (same depth) + unified AG & 54.4 & 52.8 & 69.1 & 40.4 \\
\ablStdModel \label{ablStdModel:4str+cat} & $T=4$ + concatenation & 52.2 & 50.5 & 66.4 & 37.7 \\
\ablStdModel \label{ablStdModel:4str+add} & $T=4$ + addition & 52.5 & 51.1 & 64.5 & 36.3 \\
\ablStdModel \label{ablStdModel:4str+ags} & $T=4$ + separate AGs & 53.6 & 51.0 & 68.3 & 39.4 \\
\ablStdModel \label{ablStdModel:4str+uag(streamwise)} & $T=4$ + unified AG (stream-wise) & 54.4 & 51.9 & 67.9 & 40.4 \\
\ablStdModel \label{ablStdModel:4str+fixednum} & $T=4$ + learned-and-fixed gates & 52.9 & 50.8 & 64.5 & 36.2 \\
\ablStdModel \label{ablStdModel:1str} & $T=1$ & 43.3 & 42.1 & 52.9 & 26.8 \\
\ablStdModel \label{ablStdModel:2str+uag} & $T=2$ + unified AG & 52.2 & 50.0 & 65.6 & 37.0 \\
\ablStdModel \label{ablStdModel:3str+uag} & $T=3$ + unified AG & 54.0 & 51.8 & 67.7 & 39.6 \\
\bottomrule
\end{tabular}
\end{table}

\keypoint{Ablation Study}
Table~\ref{tab:ablation_omni_scale} evaluates our architectural design choices for omni-scale feature learning. The primary model is model~\ref{ablStdModel:4str+uag}. $T$ denotes the stream cardinality in Eq.~\eqref{eq:mult_stream_residual}. The results are summarised as follows.
\textbf{(1)} \textit{vs.~standard convolutions:}
Overall, factorising convolutions does not significantly harm the performance while has a positive effect on large datasets like MSMT17 (model~\ref{ablStdModel:4str(fullconv)+uag} vs.~\ref{ablStdModel:4str+uag}). This means our design helps maintain the representational power while shrinking the model size by more than $3\! \times$.
\textbf{(2)} \textit{vs.~ResNeXt-like design:}
OSNet is transformed into a ResNeXt-like architecture by making all streams homogeneous in depth while preserving the unified AG, which refers to model~\ref{ablStdModel:4str(t=1)+uag}. We observe that this variant is clearly outperformed by the primary model, which further validates the necessity of the omni-scale design.
\textbf{(3)} \textit{Multi-scale fusion strategy:}
We change the way how features of different scales are aggregated. The baselines are concatenation (model~\ref{ablStdModel:4str+cat}) and addition (model~\ref{ablStdModel:4str+add}). The primary model is clearly better than the two baselines. Nevertheless, models~\ref{ablStdModel:4str+cat} and \ref{ablStdModel:4str+add} are still much better than the single-scale architecture (model~\ref{ablStdModel:1str}).
\textbf{(4)} \textit{Unified AG vs.~separate AGs:}
When separate AGs are learned for each feature stream, the model size is increased and the nice property in gradient computation (Eq.~\eqref{eq:grad_ag}) vanishes. Empirically, unifying AG improves the performance (model~\ref{ablStdModel:4str+uag} vs.~\ref{ablStdModel:4str+ags}), despite having less parameters.
\textbf{(5)} \textit{Channel-wise gates vs.~stream-wise gates:}
By turning the channel-wise gates into stream-wise gates (model~\ref{ablStdModel:4str+uag(streamwise)}), the performance deteriorates. As feature channels represent numerous visual concepts and encapsulate sophisticated correlations~\cite{fong2018net2vec}, it is advantageous to use channel-specific gates.
\textbf{(6)} \textit{Dynamic gates vs.~static gates:}
In model~\ref{ablStdModel:4str+fixednum}, feature streams are fused by static (learned-and-then-fixed) channel-wise gates to mimic the design in~\cite{qian2017multi}. As a result, the performance drops noticeably compared with that of dynamic gating (model~\ref{ablStdModel:4str+uag}). Therefore, adapting the scale fusion for individual input images is essential.
\textbf{(7)} \textit{Evaluation on stream cardinality:}
The results are substantially boosted from $T\!=\!1$ (model~\ref{ablStdModel:1str}) to $T\!=\!2$ (model~\ref{ablStdModel:2str+uag}), and gradually progress to $T\!=\!4$ (model~\ref{ablStdModel:4str+uag}).

\keypoint{Model Shrinking Hyperparameters}
We can trade-off between model size, computations and performance by adjusting the width multiplier $\beta$ and the image resolution multiplier $\gamma$. Table~\ref{tab:model_shrink_params} shows that by keeping one multiplier fixed and shrinking the other, the R1 drops off \emph{smoothly}. It is worth noting that 92.2\% R1 accuracy is obtained by a much shrunken version of OSNet with \emph{merely 0.2M parameters and 82.3M mult-adds} ($\beta\!=\!0.25$). Compared with the results in Table~\ref{tab:same_domain_big}, we can see that the shrunken OSNet is still very competitive against the latest models (most being $100\! \times$ bigger in size). This indicates that OSNet is a superior fit for efficient deployment in resource-constrained devices such as surveillance cameras with AI processors.

\begin{table}[t]
\setlength{\tabcolsep}{8.5pt}
\renewcommand{\arraystretch}{\tableCellHeight}
\centering
\footnotesize
\caption{Results of varying width multiplier $\beta$ and resolution multiplier $\gamma$ for OSNet. For input size, $\gamma=0.75$: $192\times96$; $\gamma=0.5$: $128\times64$; $\gamma=0.25$: $64\times32$.}
\label{tab:model_shrink_params}
\begin{tabular}{c|c|c|c|cc}
\toprule
\multirow{2}{*}{$\beta$} & \multirow{2}{*}{\# params} & \multirow{2}{*}{$\gamma$} & \multirow{2}{*}{Mult-Adds} & \multicolumn{2}{c}{Market1501 } \\
 & & & & R1 & mAP \\
\midrule
1.0 & 2.2M & 1.0 & 978.9M & 94.8 & 86.7 \\
0.75 & 1.3M & 1.0 & 571.8M & 94.5 & 84.1 \\
0.5 & 0.6M & 1.0 & 272.9M & 93.4 & 82.6 \\
0.25 & 0.2M & 1.0 & 82.3M & 92.2 & 77.8 \\
\midrule
1.0 & 2.2M & 0.75 & 550.7M & 94.4 & 83.7 \\
1.0 & 2.2M & 0.5 & 244.9M & 92.0 & 80.3 \\
1.0 & 2.2M & 0.25 & 61.5M & 86.9 & 67.3 \\
\midrule
0.75 & 1.3M & 0.75 & 321.7M & 94.3 & 82.4 \\
0.75 & 1.3M & 0.5 & 143.1M & 92.9 & 79.5 \\
0.75 & 1.3M & 0.25 & 35.9M & 85.4 & 65.5 \\
\midrule
0.5 & 0.6M & 0.75 & 153.6M & 92.9 & 80.8 \\
0.5 & 0.6M & 0.5 & 68.3M & 91.7 & 78.5 \\
0.5 & 0.6M & 0.25 & 17.2M & 85.4 & 66.0 \\
\midrule
0.25 & 0.2M & 0.75 & 46.3M & 91.6 & 76.1 \\
0.25 & 0.2M & 0.5 & 20.6M & 88.7 & 71.8 \\
0.25 & 0.2M & 0.25 & 5.2M & 79.1 & 56.0 \\
\bottomrule
\end{tabular}
\end{table}

\begin{figure*}[t]
\centering
\includegraphics[width=0.99\textwidth]{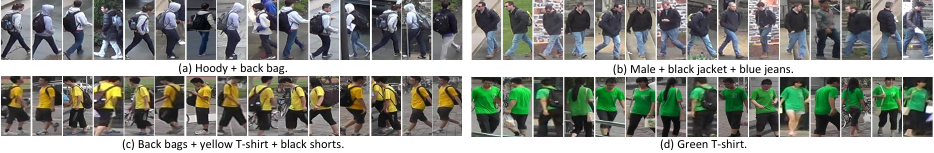}
\caption{Image clusters of similar gating vectors. The visualisation shows that the proposed unified aggregation gate is capable of learning the combination of homogeneous and heterogeneous scales conditioned on the input data.}
\label{fig:clusterimgs}
\end{figure*}

\begin{figure}[t]
\centering
\includegraphics[width=0.9\columnwidth]{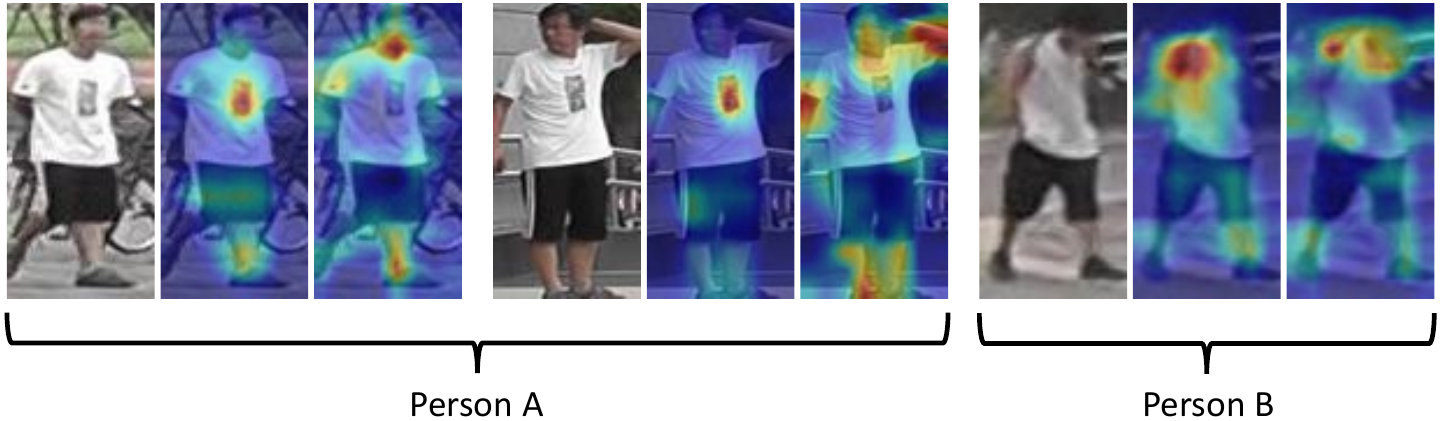}
\caption{Visual attention insight. Each triplet contains, from left to right, the original image, activation map of OSNet, and the single-scale baseline. OSNet can detect subtle differences between visually similar persons.}
\label{fig:vis_actmap}
\end{figure}

\keypoint{Visualisation of Unified Aggregation Gate}
As the gating vectors produced by the AG inherently encode the way how the omni-scale feature streams are aggregated, we can understand what the AG sub-network has learned by visualising images of similar gating vectors. To this end, we 1) concatenate the gating vectors of four streams in the last bottleneck as the representations of test images, 2) perform $k$-means clustering, and 3) select top-15 images closest to the cluster centres. Fig.~\ref{fig:clusterimgs} shows four example clusters where images within the same cluster exhibit similar patterns, i.e.~combinations of global-scale and local-scale appearance.

\keypoint{Visualisation of Learned Features}
To understand how our designs help OSNet learn discriminative features, we visualise the activations of the last convolutional feature maps to investigate where the network focuses on to extract features. Following~\cite{zagoruyko2017paying}, the activation maps are computed as the sum of absolute-valued feature maps along the channel dimension followed by a spatial $\ell_2$ normalisation. Fig.~\ref{fig:vis_actmap} compares the activation maps of OSNet and the single-scale baseline (model~\ref{ablStdModel:1str} in Table~\ref{tab:ablation_omni_scale}). It is clear that OSNet can capture the local discriminative patterns of Person A (e.g., the clothing logo) which distinguish Person A from Person B. In contrast, the single-scale model over-concentrates on the face region, which is unreliable for re-ID due to low resolution of surveillance images. This qualitative result shows that our multi-scale design and unified aggregation gate enable OSNet to identify subtle differences between visually similar persons---a vital ability for accurate re-ID.

\begin{table}[t]
\setlength{\tabcolsep}{4pt}
\renewcommand{\arraystretch}{\tableCellHeight}
\centering
\footnotesize
\caption{Ablation study for instance normalisation and architecture search.}
\label{tab:ablation_IN_NAS}
\begin{tabular}{l | c c c c | c c c c}
\toprule
\multirow{2}{*}{Method} & \multicolumn{4}{c|}{Market1501$\to$Duke} & \multicolumn{4}{c}{Duke$\to$Market1501} \\
 & R1 & R5 & R10 & mAP & R1 & R5 & R10 & mAP \\
\midrule
IBN-Net~\cite{pan2018ibn} & 43.7 & 59.1 & 65.2 & 24.3 & 50.7 & 69.1 & 76.3 & 23.5 \\
OSNet & 44.7 & 59.6 & 65.4 & 25.9 & 52.2 & 67.5 & 74.7 & 24.0 \\
OSNet-IBN & 47.9 & 62.7 & 68.2 & 27.6 & 57.8 & 74.0 & 79.5 & 27.4 \\
OSNet-AIN & \textbf{52.4} & \textbf{66.1} & \textbf{71.2} & \textbf{30.5} & \textbf{61.0} & \textbf{77.0} & \textbf{82.5} & \textbf{30.6} \\
\bottomrule
\end{tabular}
\end{table}

\begin{table}[t]
\setlength{\tabcolsep}{10pt}
\renewcommand{\arraystretch}{\tableCellHeight}
\centering
\footnotesize
\caption{Performance of OSNet-AIN in the same-domain re-ID setting.}
\label{tab:IN_same_domain_result}
\begin{tabular}{l | c c | c c}
\toprule
\multirow{2}{*}{Method} & \multicolumn{2}{c|}{Market1501} & \multicolumn{2}{c}{Duke} \\
 & R1 & mAP & R1 & mAP \\
\midrule
OSNet & 94.8 & 86.7 & 88.2 & 76.7 \\
OSNet-AIN & 94.2 & 84.4 & 87.9 & 74.2 \\
\bottomrule
\end{tabular}
\end{table}

\begin{table*}[t]
\setlength{\tabcolsep}{4.5pt}
\renewcommand{\arraystretch}{\tableCellHeight}
\centering
\footnotesize
\caption{Comparison with current state-of-the-art unsupervised domain adaptation methods in the cross-domain re-ID setting.
OSNet-AIN achieves highly comparable performance despite \emph{only using the source training data without per-domain model adaptation}. $U$: Unlabelled.}
\label{tab:xdomain_single_src}
\begin{tabular}{l | c | c | c c c c || c | c c c c}
\toprule
\multirow{2}{*}{Method} & \multirow{2}{*}{Venue} & \multirow{2}{*}{Source} & \multicolumn{4}{c||}{Target: Duke} & \multirow{2}{*}{Source} & \multicolumn{4}{c}{Target: Market1501} \\
 & & & R1 & R5 & R10 & mAP & & R1 & R5 & R10 & mAP \\
\midrule
MMFA~\cite{lin2018multi} & BMVC'18 & Market1501 + Duke ($U$) & 45.3 & 59.8 & 66.3 & 24.7 & Duke + Market1501 ($U$) & 56.7 & 75.0 & 81.8 & 27.4 \\
SPGAN~\cite{deng2018image} & CVPR'18 & Market1501 + Duke ($U$) & 46.4 & 62.3 & 68.0 & 26.2 & Duke + Market1501 ($U$) & 57.7 & 75.8 & 82.4 & 26.7 \\
TJ-AIDL~\cite{wang2018transferable} & CVPR'18 & Market1501 + Duke ($U$) & 44.3 & 59.6 & 65.0 & 23.0 & Duke + Market1501 ($U$) & 58.2 & 74.8 & 81.1 & 26.5 \\
ATNet~\cite{liu2019adaptive} & CVPR'19 & Market1501 + Duke ($U$) & 45.1 & 59.5 & 64.2 & 24.9 & Duke + Market1501 ($U$) & 55.7 & 73.2 & 79.4 & 25.6 \\
CamStyle~\cite{zhong2019camstyle} & TIP'19 & Market1501 + Duke ($U$) & 48.4 & 62.5 & 68.9 & 25.1 & Duke + Market1501 ($U$) & 58.8 & 78.2 & 84.3 & 27.4 \\
HHL~\cite{zhong2018gen} & ECCV'18 & Market1501 + Duke ($U$) & 46.9 & 61.0 & 66.7 & 27.2 & Duke + Market1501 ($U$) & 62.2 & 78.8 & 84.0 & 31.4 \\
ECN~\cite{zhong2019invariance} & CVPR'19 & Market1501 + Duke ($U$) & 63.3 & 75.8 & 80.4 & 40.4 & Duke + Market1501 ($U$) & 75.1 & 87.6 & 91.6 & 43.0 \\
SSG~\cite{fu2019self} & ICCV'19 & Market1501 + Duke ($U$) & 73.0 & 80.6 & 83.2 & 53.4 & Duke + Market1501 ($U$) & 80.0 & 90.0 & 92.4 & 58.3 \\
OSNet-AIN (\emph{ours}) & This work & Market1501 & 52.4 & 66.1 & 71.2 & 30.5 & Duke & 61.0 & 77.0 & 82.5 & 30.6 \\ 
\midrule
MAR~\cite{yu2019unsupervised} & CVPR'19 & MSMT17+Duke ($U$) & 67.1 & 79.8 & - & 48.0 & MSMT17+Market1501 ($U$) & 67.7 & 81.9 & - & 40.0 \\
PAUL~\cite{yang2019patch} & CVPR'19 & MSMT17+Duke ($U$) & 72.0 & 82.7 & 86.0 & 53.2 & MSMT17+Market1501 ($U$) & 68.5 & 82.4 & 87.4 & 40.1 \\
OSNet-AIN (\emph{ours}) & This work & MSMT17 & 71.1 & 83.3 & 86.4 & 52.7 & MSMT17 & 70.1 & 84.1 & 88.6 & 43.3 \\ 
\bottomrule
\end{tabular}
\end{table*}

\begin{table}
\setlength{\tabcolsep}{2.5pt}
\renewcommand{\arraystretch}{\tableCellHeight}
\centering
\footnotesize
\caption{Cross-domain results on the more challenging MSMT17 dataset.}
\label{tab:xdomain_test_msmt}
\begin{tabular}{l | c | c c c c}
\toprule
\multirow{2}{*}{Method} & \multirow{2}{*}{Source} & \multicolumn{4}{c}{Target: MSMT17} \\
& & R1 & R5 & R10 & mAP \\
\midrule
ECN~\cite{zhong2019invariance} & Market1501 + MSMT17 ($U$) & 25.3 & 36.3 & 42.1 & 8.5 \\
SSG~\cite{fu2019self} & Market1501 + MSMT17 ($U$) & 31.6 & - & 49.6 & 13.2 \\
OSNet-AIN (\emph{ours}) & Market1501 & 23.5 & 34.5 & 40.2 & 8.2 \\
\midrule
ECN~\cite{zhong2019invariance} & Duke + MSMT17 ($U$) & 30.2 & 41.5 & 46.8 & 10.2 \\
SSG~\cite{fu2019self} & Duke + MSMT17 ($U$) & 32.2 & - & 51.2 & 13.3 \\
OSNet-AIN (\emph{ours}) & Duke & 30.3 & 42.2 & 47.9 & 10.2 \\
\bottomrule
\end{tabular}
\end{table}

\subsection{Cross-Domain Person Re-Identification} \label{subsec:crossDomainReID}
In this section, we evaluate the domain-generalisable OSNet with IN, i.e.~OSNet-AIN, in the cross-dataset re-ID setting. In particular, we aim to assess the generalisation performance of OSNet-AIN by first training the model on a source dataset and then \emph{directly} testing its performance on an unseen target dataset \emph{without the need for per-domain model adaptation}. This differs significantly from the current state-of-the-art unsupervised domain adaptation (UDA) methods~\cite{zhong2019invariance,liu2019adaptive,yu2019unsupervised,yang2019patch}, which require per-domain adaptation on the target domain data (hence more computationally expensive and less scalable).

\keypoint{Architecture Search}
We first discuss the experimental details regarding how OSNet-AIN is searched. For the dataset to perform NAS, we choose MSMT17, which contains the largest camera network (15 cameras) and has diverse image qualities/styles (collected in four days of different weather conditions within a month). Once the network architecture is found, we directly transfer it to other re-ID datasets without re-searching. The over-parameterised network (Fig.~\ref{fig:search_space}) is trained from scratch using SGD, batch size of 512, initial learning rate of 0.1 and weight decay of 5e-4 for 120 epochs on 8 Tesla V100 32GB GPUs. The learning rate is annealed down to zero using the cosine annealing trick~\cite{cosineLR} without restart. The softmax temperature $\lambda$ (Eq.~\eqref{eq:gumbel_softmax}) starts from 10 and decreases by 0.5 every 20 epochs (the minimum is fixed to 1.\footnote{Setting $\lambda < 1$ makes the training unstable.}) Though a larger Monte Carlo sampling number $S$ in Eqs.~\eqref{eq:mc_grad_theta}~\&~\eqref{eq:mc_grad_pi} is theoretically better for convergence, we empirically found that setting $S\! =\! 1$ worked well and greatly shortened the training time. The objective function is the cross-entropy loss with label smoothing. Random flip and colour jittering are used for data augmentation.

\keypoint{Datasets and Settings}
Following the recent UDA re-ID works~\cite{zhong2019invariance,zhong2019camstyle,liu2019adaptive,yu2019unsupervised,yang2019patch}, we experiment with Market1501$\to$Duke, Duke$\to$Market1501, MSMT17$\to$Market1501/Duke, and Market1501/Duke$\to$MSMT17. When the source dataset is MSMT17, we use all 126,441 images of 4,101 identities for model training, following~\cite{yu2019unsupervised,yang2019patch}. OSNet-AIN is first pre-trained on ImageNet and then fine-tuned on a source dataset for re-ID on a target dataset. The training pipeline and details for this cross-domain setting follow those used in the same-domain setting, except that the maximum epoch is 100 for Market1501/Duke and 50 for MSMT17 and the data augmentation includes random flip and colour jittering.

\keypoint{Effect of Instance Normalisation}
Table~\ref{tab:ablation_IN_NAS} shows that OSNet-AIN significantly improves upon OSNet by 7.7\% R1 and 4.6\% mAP on Market1501$\to$Duke and 8.8\% R1 and 6.6\% mAP on Duke$\to$Market1501. This justifies the effectiveness of IN for cross-domain re-ID. Comparing OSNet with IBN-Net, we observe that our basic omni-scale backbone is already stronger than the IN-equipped ResNet50 model. This suggests that our omni-scale network design is not only effective for learning discriminative re-ID features for source datasets, but also helps the learning of generalisable person features for unseen target datasets. This is because omni-scale features capture both global- and local-scale patterns, intuitively a domain-agnostic capability.

Table~\ref{tab:IN_same_domain_result} further tests the model performance effect of IN in the same-domain re-ID setting by comparing OSNet-AIN with OSNet. We observe that IN slightly decreases the performance. This is not surprising: during feature learning IN progressively removes dataset-specific features that are detrimental to cross-domain re-ID but potentially beneficial to same-domain recognition. This is thus a price one has to pay to make the model more generalisable to unseen domains. Note that the performance of OSNet-AIN in the same-domain setting is still very competitive when compared with the state-of-the-art alternatives in Table~\ref{tab:same_domain_big}.

\keypoint{Search vs.~Engineering}
To justify the contribution of our architecture search algorithm, we hand-engineer an OSNet+IN model by mimicking the design rule in IBN-Net. Specifically, we add IN only to the lowest layers in OSNet (conv1 \& conv2 as shown in Table~\ref{tab:netarch}), which we call \emph{OSNet-IBN}. Table~\ref{tab:ablation_IN_NAS} shows that OSNet-AIN significantly outperforms OSNet-IBN by 4.5\% R1 and 2.9\% mAP on Market1501$\to$Duke and 3.2\% R1 and 3.2\% mAP on Duke$\to$Market1501. This strongly demonstrates the superiority of our architecture search algorithm over handcrafted architecture design. Given that the search space for the entire network contains $4^6\!=\!4096$ different configurations, it is much more efficient to learn the network configuration rather than exhaustively trying all possible choices.

\keypoint{Comparative Results}
Table~\ref{tab:xdomain_single_src} compares OSNet-AIN with current state-of-the-art cross-domain re-ID methods based on UDA, which obtain unfair gains by using the target data. It is clear that OSNet-AIN achieves promising results on the target datasets despite only using source data---it outperforms most UDA methods in terms of R1. When the source dataset is large such as MSMT17, OSNet-AIN substantially improves the cross-domain performance (R1) from 52.4\% to 71.1\% on Duke and 61.0\% to 70.1\% on Market1501, which are close to the performance of the latest UDA methods. It is noted that most top-performing UDA methods (ECN, HHL, CamStyle and ATNet in the small source case) rely on image-to-image translation models such as CycleGAN~\cite{CycleGAN} to synthesise target-style images and have complex adaptation procedures for each target domain. These thus severely hinder their deployment in real-world applications where out-of-the-box solution is desired. In contrast, OSNet-AIN enables adaptation-free, plug-and-play deployment once trained on a source dataset. Notably, in the more challenging scenario where the source dataset is small but the target dataset is large, i.e.~Market1501$\to$MSMT17, as shown in Table~\ref{tab:xdomain_test_msmt}, OSNet-AIN can achieve performance on par with ECN---the latter benefits from more unlabelled target data from MSMT17.

To further demonstrate our model's capability, we conduct evaluation on multi-source domain generalisation~\cite{zhao2021learning} where a model is trained using multiple source datasets, such as a combination of Market1501, Duke, and CUHK03, and tested on an unseen target dataset, such as MSMT17. Please see Appendix~\ref{appx:msdg} and Table~\ref{tab:msdg}.

\section{Conclusion}
In this paper, we presented OSNet, a lightweight CNN architecture that is capable of learning omni-scale feature representations for person re-ID. Compared with existing re-ID CNNs, OSNet has the unique ability to learn multi-scale features explicitly inside each building block, where the unified aggregation gate dynamically fuses multi-scale features to produce omni-scale features. To improve cross-domain generalisation, we equipped OSNet with instance normalisation via differentiable architecture search, resulting in a domain-adaptive variant called OSNet-AIN. In the same-domain re-ID setting, the results showed that OSNet achieves state-of-the-art performance while being much smaller than ResNet-based opponents. In the cross-domain re-ID setting, OSNet-AIN exhibited a remarkable generalisation ability on unseen target datasets, beating most recent UDA methods without using per-domain model adaptation on target domain data.


%

\appendices
\section{Multi-Source Domain Generalisation} \label{appx:msdg}
To further demonstrate the effectiveness of OSNet-AIN, we conduct experiments in the multi-source domain generalisation setting following~\cite{zhao2021learning},\footnote{\url{https://github.com/HeliosZhao/M3L.}} i.e.~a model is trained using multiple source datasets rather than a single dataset. The aim is to show that integrating instance normalisation into OSNet via architecture search is better than hand-engineering.

\keypoint{Datasets}
Following~\cite{zhao2021learning}, we use the four largest re-ID datasets, namely Market1501, Duke, CUHK03,\footnote{The 767/700 split is used for both training and testing.} and MSMT17. In particular, three datasets are used for training and the remaining one is used for testing. During model training, we only use the training split of the source datasets.

\keypoint{Training Details}
All models are trained using the cross-entropy loss with label smoothing. We simply build a single large classification layer for classifying all identities combined from different datasets. We keep the training parameters the same as those used in Sec.~\ref{subsec:crossDomainReID}. The details of the training parameters can be found in our Github repository.\footnote{Please see {im\_osnet\_ain\_x1\_0\_softmax\_256x128\_amsgrad\_cosine.yaml}. The maximum epoch is set to 50 instead of 100.}

\keypoint{Results}
The results are presented in Table~\ref{tab:msdg}. Comparing OSNet with OSNet-IBN, we observe that using instance normalisation layers improves the plain OSNet's performance, especially on the most challenging test domain (MSMT17). By automating the architecture engineering process for integrating instance normalisation layers, the performance is further improved (OSNet-AIN vs.~OSNet-IBN).

\begin{table*}[t]
\renewcommand{\arraystretch}{\tableCellHeight}
\setlength{\tabcolsep}{10pt}
\centering
\footnotesize
\caption{Results in the multi-source domain generalisation setting using MSMT17 (MS), Market1501 (M), Duke (D), and CUHK03 (C).}
\label{tab:msdg}
\begin{tabular}{l | cc | cc | cc | cc}
\toprule
\multirow{2}{*}{Model} & \multicolumn{2}{c|}{MS+D+C$\to$M} & \multicolumn{2}{c|}{MS+M+C$\to$D} & \multicolumn{2}{c|}{MS+D+M$\to$C} & \multicolumn{2}{c}{D+M+C$\to$MS} \\
& mAP & R1 & mAP & R1 & mAP & R1 & mAP & R1 \\
\midrule
OSNet & 44.2 & 72.5 & 47.0 & 65.2 & 23.3 & 23.9 & 12.6 & 33.2 \\
OSNet-IBN & 44.9 & 73.0 & 45.7 & 64.6 & 25.4 & 25.7 & 16.2 & 39.8 \\
OSNet-AIN & 45.8 & 73.3 & 47.2 & 65.6 & 27.1 & 27.4 & 16.2 & 40.2 \\
\bottomrule
\end{tabular}
\end{table*}





\ifCLASSOPTIONcaptionsoff
  \newpage
\fi


{
\bibliographystyle{IEEEtran}
\bibliography{reference}
}

\end{document}